\pdfoutput=1

\documentclass[11pt]{article}

\usepackage[final]{acl}

\usepackage{times}
\usepackage{latexsym}
\usepackage{natbib}
\usepackage{amsfonts}
\usepackage[T1]{fontenc}
\usepackage{graphicx}
\usepackage{xspace}
\newcommand{\methodnameshort}{\textsc{ICG}\xspace}
\usepackage[utf8]{inputenc}

\usepackage{microtype}
\usepackage{pifont}
\usepackage{inconsolata}
\usepackage{float}

\usepackage{amsmath}
\usepackage{enumitem}
\usepackage{booktabs}
\usepackage{amsthm}
\usepackage{balance}
\usepackage{soul}
\usepackage{url}
\usepackage[utf8]{inputenc}

\title{ICG: Improving Cover Image Generation via MLLM-based Prompting and Personalized Preference Alignment}

\author{
\textbf{Zhipeng Bian\textsuperscript{1,2}},~
\textbf{Jieming Zhu\textsuperscript{2}$\footnotemark[1]$},~
\textbf{Qijiong Liu\textsuperscript{3}},~
\textbf{Wang Lin\textsuperscript{4}},~
\textbf{Guohao Cai\textsuperscript{2}}, \\
\textbf{Zhaocheng Du\textsuperscript{2}},~
\textbf{Jiacheng Sun\textsuperscript{2}},~
\textbf{Zhou Zhao\textsuperscript{4}},~
\textbf{Zhenhua Dong\textsuperscript{2}} \\
\textsuperscript{1}Huazhong University of Science and Technology \quad \textsuperscript{2}Huawei Noah's Ark Lab \\
\textsuperscript{3}Hong Kong Polytechnic University \quad \textsuperscript{4}Zhejiang University \\
\texttt{bian\_zhipeng@hust.edu.cn} \quad \texttt{jiemingzhu@ieee.org} \quad \texttt{liu@qijiong.work} \\
\texttt{\{caiguohao,duzhaocheng,sunjiacheng,dongzhenhua\}@huawei.com} \\ \texttt{\{linwanglw,zhaozhou\}@zju.edu.cn} 
}

\begin{document}
\maketitle
\renewcommand{\thefootnote}{\fnsymbol{footnote}}
\footnotetext[1]{\ Corresponding Author.}
\begin{abstract}
Recent advances in multimodal large language models (MLLMs) and diffusion models (DMs) have opened new possibilities for AI-generated content. Yet, personalized cover image generation remains underexplored, despite its critical role in boosting user engagement on digital platforms. We propose \methodnameshort, a novel framework that integrates MLLM-based prompting with personalized preference alignment to generate high-quality, contextually relevant covers. \methodnameshort extracts semantic features from item titles and reference images via meta tokens, refines them with user embeddings, and injects the resulting personalized context into the diffusion model. To address the lack of labeled supervision, we adopt a multi-reward learning strategy that combines public aesthetic and relevance rewards with a personalized preference model trained from user behavior. Unlike prior pipelines relying on handcrafted prompts and disjointed modules, \methodnameshort employs an adapter to bridge MLLMs and diffusion models for end-to-end training. Experiments demonstrate that \methodnameshort significantly improves image quality, semantic fidelity, and personalization, leading to stronger user appeal and offline recommendation accuracy in downstream tasks. As a plug-and-play adapter bridging MLLMs and diffusion models, \methodnameshort is compatible with common checkpoints and requires no ground-truth labels during optimization.
\end{abstract}

\section{Introduction}
Large language models (LLMs) and diffusion models (DMs) have driven the rise of AI-generated content (AIGC) in applications such as personal assistants, chatbots, digital art, and cover image generation~\cite{CoverGeneration,MusicCover,AdCover}. In recommender systems—especially news feeds—blurry, mismatched, or unappealing covers are common, undermining user engagement. Thus, improving cover image generation is critical to enhancing recommendation quality.


\label{introduction}
\begin{figure}
    \centering
    \includegraphics[width=\linewidth]{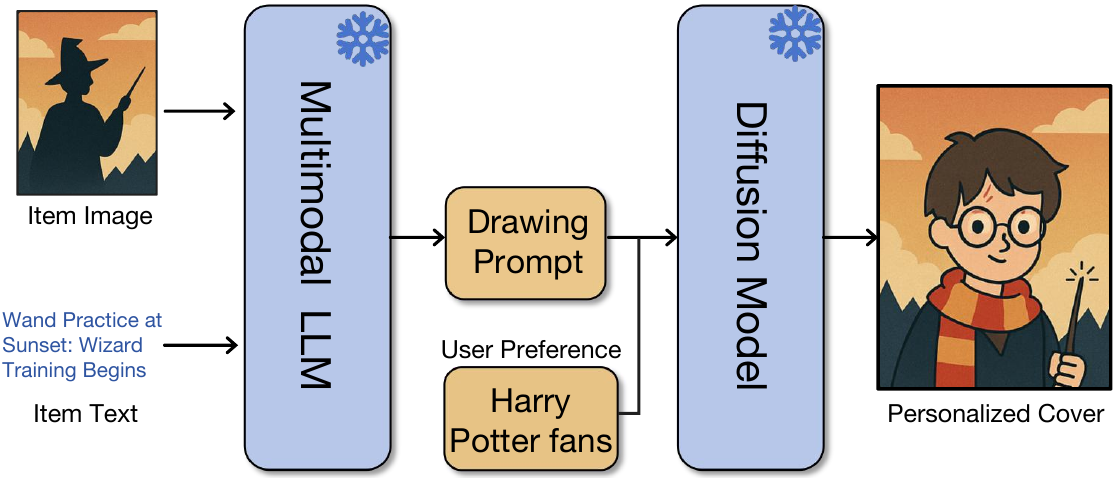}
    \caption{The overall pipeline for cover image generation.}
    \label{fig:baseline}
    \vspace{-0.3cm}
\end{figure}
Text-to-image models such as Stable Diffusion~\cite{RombachBLEO22}, Midjourney~\cite{midjourney}, and DALLE-3~\cite{betker2023improving} are widely used by designers and publishers for banner and cover image generation. However, they rely heavily on manually crafted prompts and careful prompt engineering, which limits scalability for platforms handling millions of items, such as news aggregators, streaming services (e.g., Netflix, YouTube), and social media feeds (e.g., TikTok, Instagram). In these scenarios, visually appealing and context-relevant cover images are critical for capturing user attention and improving engagement. As shown in Figure~\ref{fig:baseline}, a promising solution is to use multimodal large language models (MLLMs) to automatically extract semantics from raw item content and generate prompts for Stable Diffusion. Despite its simplicity, this pipeline faces several challenges in practical adoption.

Firstly, although MLLM-based prompt generation eliminates the need to manually craft prompts for each item, it still requires careful design of prompt instructions for the MLLMs. Prior works such as BeautifulPrompt~\cite{BeautifulPrompt}, Promptist~\cite{PromptOptimizer}, and UF-FGTG~\cite{UF-FGTG} aim to automate or refine prompts using large language models, but they focus on improving existing prompt text. In contrast, our task starts from raw item content (e.g., titles), rendering these methods inapplicable. Furthermore, the absence of golden prompt references for cover images limits the possibility of supervised fine-tuning for MLLMs in this setting.

Secondly, the current pipeline is disjointed and lacks end-to-end optimization, leading to issues such as MLLM hallucinations and misalignment with diffusion models, which often result in low-quality or semantically irrelevant covers. This hinders error correction and model refinement. Recent progress in multimodal AI has produced models like MiniGPT-5~\cite{MiniGPT5}, SEED-LLaMA~\cite{GeZZGLWS24}, and Kosmos-G~\cite{Kosmos-G}, which integrate MLLMs with diffusion decoders for unified understanding and generation. However, they still rely heavily on user-crafted prompts. In addition, the scarcity of high-quality cover images limits supervision when generating directly from raw item content.

Thirdly, current text-to-image generation methods lack personalization, often producing covers that fail to reflect user preferences and reduce engagement. For instance, male users may prefer dark, professional styles, while female users may favor pink, cute designs. Aligning covers with individual tastes can boost click-through rates and user experience. Prior work like PMG~\cite{PMG} and DiFashion~\cite{PersonalOutfit} explores this direction but has key limitations: (1) Both use the next item’s image as the training target, assuming high-quality covers—often untrue in practice; (2) PMG represents preferences as discrete keywords via LLMs, hindering end-to-end optimization and fine-grained preference capture. Consequently, these methods often fail to generate visually appealing, truly personalized outputs.

To address these challenges, we propose \methodnameshort, a unified framework for personalized cover generation that integrates MLLMs with reward-based optimization. It leverages item content—comprising a reference image and title—to retain original semantics, while personalization is guided by user interaction histories. Textual and visual inputs are encoded by MLLMs, with meta tokens capturing contextual features that are injected into the diffusion model via an adapter for end-to-end training. User features are fused with context to condition generation. The model is optimized using a differentiable multi-reward framework, combining public aesthetic and relevance scores with a personalized reward model trained on user-item interactions, enabling content-aligned and user-specific generation.

The main contributions of this work are:

(1) We present the first framework that integrates MLLMs with reward learning for personalized cover image generation, demonstrating its effectiveness in recommendation scenarios.

(2) We introduce meta tokens to capture contextual semantics and fuse them with user embeddings via a plug-and-play adapter into a diffusion model. A multi-reward learning framework enables end-to-end training guided by aesthetics, content relevance, and user preference alignment—without requiring explicit supervision.

(3) Extensive experiments show that \methodnameshort consistently outperforms prior methods in aesthetics, semantic fidelity, and personalization, leading to improved user engagement.

\begin{figure*}
    \centering
    \begin{minipage}[b]{\linewidth}
        \centering
        \centerline{\includegraphics[width=\linewidth]{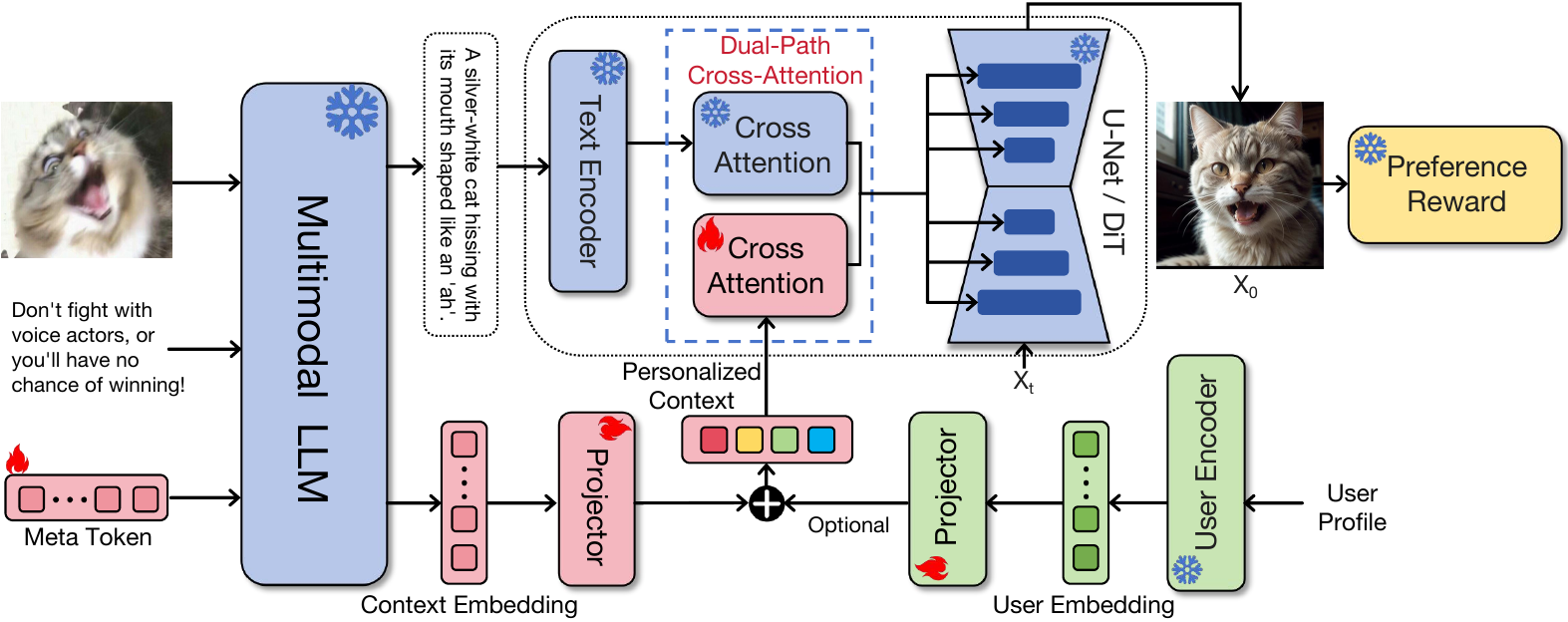}}
    \end{minipage}
    \caption{Overview of the proposed method. The model takes a reference image, title, and meta token to generate context embeddings via a Multimodal LLM. Combined with user embeddings, personalized features are injected into a diffusion model through a dual-path adapter. Reward models evaluate the output and guide training via feedback.}
    \label{fig:overview}
\end{figure*}

\section{Related Work}

\subsection{Conditional Image Generation}
Conditional image generation enables personalized synthesis from inputs like text, poses, edges, semantic maps, and reference images. Text-guided models such as CLIP encode semantics into latent space. Diffusion models like Stable Diffusion~\cite{RombachBLEO22} set the current standard. Methods like ControlNet~\cite{ZhangRA23} and MoMA~\cite{MoMA} enhance generation with structured control. For personalization, user behavior-based conditioning has been explored. DiFashion~\cite{PersonalOutfit} uses interaction history but assumes high-quality inputs; CG4CTR~\cite{AdCover} applies reward filtering but lacks end-to-end learning. Both focus on specific domains, whereas our method targets general-purpose cover generation and is thus not directly comparable.

\subsection{Automated Assessment of Image Generation}
Traditional metrics such as IS~\cite{SalimansGZCRCC16}, FID~\cite{HeuselRUNH17}, and CLIP Score~\cite{radford2021learning} are widely used to assess image fidelity and text–image consistency, but they fail to capture subjective human preferences. 
To bridge this gap, several preference-aligned evaluation models have been proposed, including PickScore~\cite{Pick_a_Pic}, HPSv2~\cite{HPSv2}, ImageReward~\cite{ImageReward}, and the Multi-dimensional Preference Score (MPS)~\cite{zhang2024learning}. 
These methods fine-tune vision–language models on large-scale human-labeled comparisons, thereby producing scores that better reflect aesthetic appeal or semantic alignment. 
However, they primarily model \emph{general human preferences}, rather than the \emph{user-specific preferences} that are crucial for personalized generation. 

In our framework, we adopt a subset of these preference models—specifically HPSv2 and PickScore—as auxiliary training signals. 
We select them for their complementary strengths: HPSv2 captures aesthetics aligned with human judgments and PickScore provides robust comparisons for text–image relevance. 
Together, they form a diverse reward set that improves overall quality and personalization, while our personalized reward module further extends beyond these general signals to incorporate user-specific feedback.

\subsection{Multimodal Large Language Models}
Multimodal Large Language Models (MLLMs)~\cite{GPT4} extend LLMs to visual inputs via modality-specific encoders and projection layers. Recent studies~\cite{KohFS23, MiniGPT5, pan2023kosmos} explore three paradigms for image generation: (1) symbolic prompts~\cite{xia2023llmga}, (2) continuous visual features~\cite{li2024blip}, and (3) discrete tokens~\cite{GeZZGLWS24} decoded by VQ-GAN~\cite{esser2021taming} or Stable Diffusion. We adopt the continuous approach for its semantic richness and compatibility with diffusion models. More recently, unified understanding–generation models, such as ILLUME~\cite{wang2024illume}, MetaQuery~\cite{pan2025transfer}, and Janus-flow~\cite{ma2025janusflow}, have emerged to jointly perform vision–language reasoning and generation within a single architecture. These models demonstrate strong zero-shot capabilities and deliver impressive quality for generic content creation. However, they typically lack mechanisms for fine-grained user conditioning or preference-aware reward optimization, which are central to personalized cover generation. Our framework is therefore complementary: instead of competing with unified models on general-purpose tasks, we focus on injecting explicit user embeddings and optimizing with multi-reward supervision to achieve personalization-aware outputs, while viewing integration with unified architectures as a promising direction for future research.

\section{Methodology}
\label{method}
We propose \methodnameshort (Figure~\ref{fig:overview}), a framework for generating personalized cover images for short videos and movies based on user preferences. It consists of four key components: (1) MLLM-based context prompting, which extracts features from the reference image and title; (2) personalized prompting, which encodes user profiles and integrates them with context features; (3) context adaptation, which injects the personalized prompt into the diffusion model; and (4) preference alignment learning, which leverages multiple reward models—including a custom personalized reward—for supervision.

\subsection{MLLM-based Context Prompting}
\begin{figure}[t]
    \includegraphics[width=\linewidth]{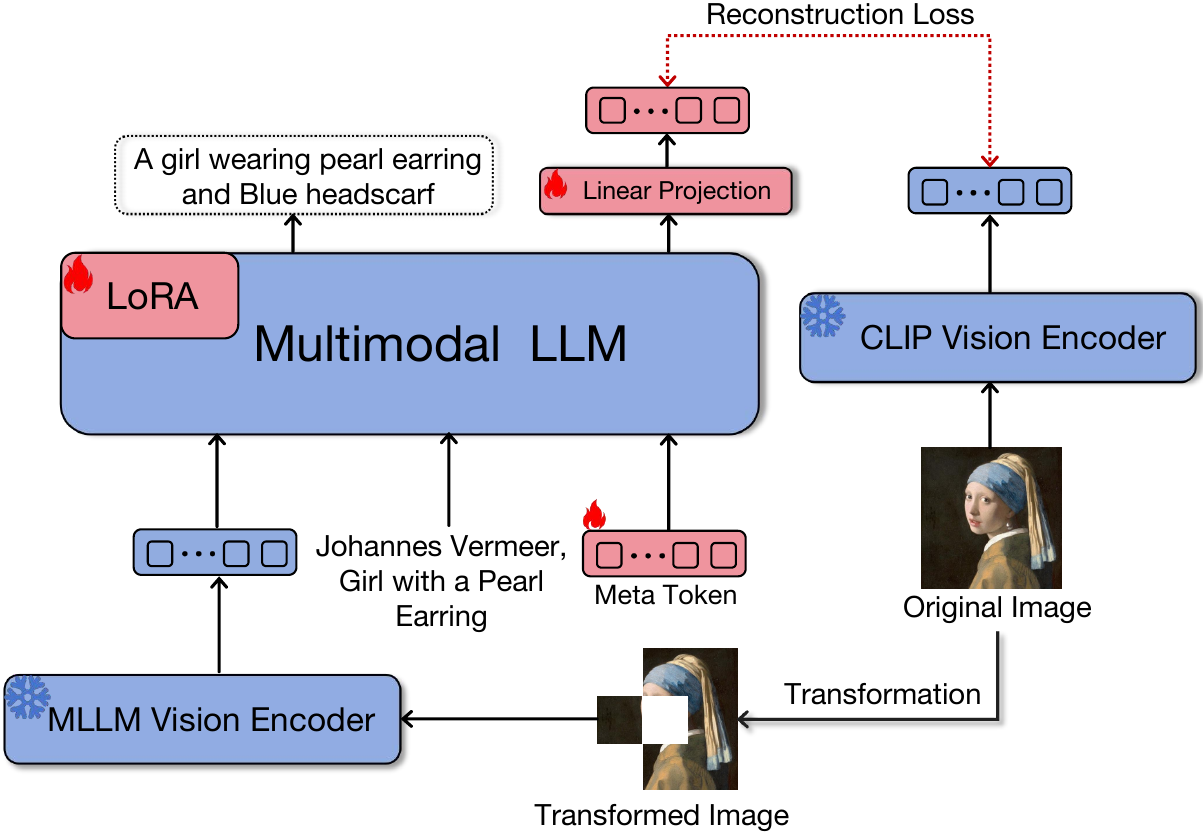}
    \caption{Model designed to train meta token.} \label{fig:prompt-tuning}
\end{figure}
We propose a Multimodal Prompt Generator based on the pre-trained MLLM Qwen2.5VL-7B~\cite{bai2025qwen2} to produce effective prompts for cover generation. The model integrates visual cues from a reference image ($I_{ref}$) and textual content ($T_{title}$), guided by a system instruction ($T_{sys}$) requesting:
\textbf{\emph{"Please generate a drawing prompt that aligns with the semantics of the specified reference cover and content title."}} This yields an explicit prompt:
\begin{equation}
\nonumber
\label{eq:explicit prompt}
\begin{aligned}
  P_{exp} = MLLM(I_{ref}, T_{title}, T_{sys}).
\end{aligned}
\end{equation}
The explicit prompt captures key entities from both modalities, ensuring basic semantic alignment. However, natural language, as a discrete representation, limits expressiveness. To address this, we introduce a meta token block that complements the prompt by capturing fine-grained multimodal context features in continuous space.

To enhance domain-specific understanding, we further design a Multimodal Generative Learning Stage (Figure~\ref{fig:prompt-tuning}). The MLLM receives $I_{ref}$, $T_{title}$, and meta tokens~\cite{KohFS23}, which are jointly attended to by text and image tokens. The meta tokens are optimized to approximate the CLIP-encoded embedding of $I_{ref}$ using a reconstruction loss:
\begin{equation}
\label{eq:prompt-tuning}
\nonumber
\begin{aligned}
    \mathcal{L} = ||MLLM(V_{enc}(I^{trans}_{ref}),T_{title},Meta\;Token)\\
    - CLIP(I_{ref})||_2^2.
\end{aligned}
\end{equation}
where $V_{enc}$ is the MLLM vision encoder, and $I^{trans}_{ref}$ is a transformed version of $I_{ref}$ (e.g., via masking, blurring, or cropping) to enhance robustness. Although CLIP embeddings alone provide strong semantic signals, our reconstruction training enables the MLLM to jointly encode textual context and transformed visual features, enriching semantic alignment and robustness beyond CLIP's single-modal representation. While meta tokens are trained with $\mathcal{L}_{rec}$, other tokens follow a standard next-token prediction objective. Once trained, the MLLM outputs prompt-contextualized embeddings for personalized cover generation.

\subsection{User-Profile-based Personalized Prompting}
The context representations and explicit text generated by the MLLM are generic and lack personalization, limiting their ability to reflect diverse user preferences. To address this, we introduce a \textbf{User-Profile-Based Personalized Prompt Generator}, which encodes user attributes—such as gender, age, occupation, and preferred cover types—as personalized style preferences to guide visual output. For example, a 27-year-old male teacher favoring cartoons and children's movies would receive prompts adapted to cartoon-style aesthetics.

Formally, the multimodal context features ($C_{ref}$) are projected into $N_c$ hidden embeddings via a linear layer. In parallel, user embeddings ($U_{pre}$), obtained from a pretrained user encoder (e.g., a two-tower CTR model~\cite{YoutubeDNN}), are projected into $N_u$ embeddings. The two sets are concatenated one-to-one to form the final personalized context prompt $C^{per}_{ref}$:
\begin{equation}
\label{eq:projection}
\nonumber
\begin{aligned}
    &Proj = LayerNorm\left(Linear(*)\right),\\
    C^{per}_{ref} &= \;Concat\left(Proj(C_{ref}),Proj(U_{pre})\right).
\end{aligned}
\end{equation}
This provides a unified representation for generating covers that are both semantically aligned and user-specific.
\subsection{Personalized Context Adaptation}
To inject personalized features into the pretrained diffusion model, we adopt a dual-path cross-attention mechanism inspired by Stable Diffusion~\cite{RombachBLEO22} and DiT~\cite{peebles2023scalable}, where text features are integrated into U-Net or transformer blocks via attention layers.

In each cross-attention layer, we introduce an additional branch for the personalized context. The outputs from both text and personalized paths are aggregated to capture general semantics and user-specific preferences. Given query features $Z$, text features $c_t$, and personalized features $c_p$, the updated output is:
\begin{equation}
\nonumber
\begin{aligned}
\mathbf{Z}^{new}=\text{Attention}(\mathbf{Q},\mathbf{K}^t,\mathbf{V}^t) + \text{Attention}(\mathbf{Q},\\\mathbf{K}^p,
\mathbf{V}^p).
\end{aligned}
\end{equation}
where $\mathbf{Q} = \mathbf{Z}\mathbf{W}_q$ is the query matrix. $\mathbf{K}^t$, $\mathbf{V}^t$ and $\mathbf{K}^p$, $\mathbf{V}^p$ are key-value pairs derived from $c_t$ and $c_p$, respectively. While $\mathbf{W}_q$, $\mathbf{W}^t_k$, and $\mathbf{W}^t_v$ are inherited from the original model, $\mathbf{W}^p_k$ and $\mathbf{W}^p_v$ are newly introduced and trained for personalization. To preserve the pretrained model, we freeze all original parameters and train only the newly added projection layers. This lightweight adaptation improves personalization while preserving the generalization ability of pretrained diffusion models. As illustrated in Figure~\ref{fig:overview}, user conditions are optional: context embeddings enhance generation quality, while user embeddings enable personalization when available.

\subsection{Personalized Preference Alignment Learning}
\label{sec:training with rewards}
As real personalized covers are unavailable as ground truth, traditional supervision (e.g., MSE) is not applicable. Inspired by reward learning from human feedback (RLHF), we guide training with multiple reward models. Public reward models~\cite{deng2024prdp,wallace2024diffusion} capture general aesthetics but overlook user-specific preferences. To address this, we introduce a personalized preference reward model that provides user-aware feedback, enabling joint optimization through a strategy we term Personalized Preference Alignment Learning.

\subsubsection{\textbf{Training of Personalized Preference Reward Model.}}
Following prior work, we formulate user preferences as pairwise comparisons. Users with fewer than six interactions are filtered out. For the remaining users, interacted items are ranked by relevance signals (e.g., clicks or ratings). The top $k_1$ items are labeled as positive and the bottom $k_2$ as negative, forming up to $k_1 \times k_2$ training pairs.

The reward model is built on CLIP, enhanced with transformer layers and fully connected (FC) heads. Each input includes a title, caption (generated via CLIP-Interrogator\footnote{\url{https://github.com/pharmapsychotic/clip-interrogator}}), user profile, and image. These inputs are encoded and projected as follows:
\begin{equation}
\label{eq:projection}
\nonumber
\begin{aligned}
    &\;\;\;t = CLIP_{txt}(title)\;,\;c = CLIP_{txt}(caption),\\
    &\;\;\;i = CLIP_{vis}(image)\;, \;u = CLIP_{txt}(user),\\
    &\;\;\;\;\;\;\;\;\;\;\;\;\;\;\;t_{f}= FC_{t}(t)\;,\; c_{f}= FC_{c}(c), \;\\
    &\;\;\;\;\;\;\;\;\;\;\;\;\;\;\;i_{f}= FC_{i}(i)\;,\;u_{f}= FC_{u}(u),\\
    &t_{t},i_{t},u_{t}= Transformer(concat(t_{f},c_{f}),i_{f},u_{f}),\\
    &\;\;\;\;\;\;\;\;\;\;\;\;\;\;\;p=FC_{per}(concat(t_{t},i_{t},u_{t})).
\end{aligned}
\end{equation}
where $CLIP_{txt}$ and $CLIP_{vis}$ are the CLIP text and image encoders, and $p$ is the predicted personalized preference score. The loss is defined as:
\begin{equation} \label{eq:loss}
\nonumber
\begin{split}
    \mathcal{L} = -\;\mathbb{E}_{U \sim \mathcal{D}}[\;\mathop{\log}\;(\sigma(\;p_{m} - p_{n}\;))\;]\;.
\end{split}
\end{equation}
where $p_m$ and $p_n$ are scores for more- and less-preferred items, respectively. To prevent overfitting, only the last few layers of CLIP and the added modules are trained.

\subsubsection{\textbf{Training with Multi-Reward Feedback.}}
Our goal is to generate covers that are both aesthetically appealing and aligned with user preferences. To achieve this, we employ three reward models: 1) \textbf{HPSv2}: Evaluating color vividness and content completeness; 2) \textbf{PickScore}: Measuring overall visual aesthetics; 3) \textbf{Personalized Reward Model}: Capturing user-specific preferences. Training consists of two stages: 1) \textbf{Initialization}: We align personalized features with the diffusion model using a weak CLIP-based reconstruction loss between generated images and their captions; 2) \textbf{Reward Feedback Learning}: For each sample $(title_i, ref\_img_i, caption_i)$, we extract personalized features using the multimodal LLM and user encoder. A latent $x_t$ is sampled from Gaussian noise and denoised into image $x_0$ via the diffusion model. The generated image is evaluated by all reward models. The final training objective is a weighted sum of reward losses:
\begin{equation}
\nonumber
\label{eq:projection}
\begin{aligned}
    \mathcal{L}_{\text{total}} = \lambda_h \mathcal{L}_h + \lambda_{per} \mathcal{L}_{per} + \lambda_p \mathcal{L}_p + \lambda_r \mathcal{L}_{\text{rec}}\;.
\end{aligned}
\end{equation}
where $\mathcal{L}_h$, $\mathcal{L}_{per}$, and $\mathcal{L}_p$ denote losses from HPSv2, the personalized reward model, and PickScore, respectively. $\mathcal{L}_{rec}$ ensures alignment between image and caption. All weights $\lambda$ are set to 0.25. Only the adapter and projector layers are updated, enabling efficient optimization of both personalization and visual quality.
\begin{table*}[t]
\centering
\caption{Quantitative comparisons. The best results are in \textbf{bold} and the second-best results are \underline{underlined}.}
\fontsize{8}{11}\selectfont
\begin{tabular}{lccccccccc}
\toprule
Dataset             & \multicolumn{4}{c}{PixelRec}  & \multicolumn{4}{c}{MovieLens}        \\ \cmidrule(lr){2-5}\cmidrule(lr){6-9} 
Metric & LPIPS($\downarrow$) & SSIM($\uparrow$) & FID($\downarrow$) & Aesthetics($\uparrow$) & LPIPS($\downarrow$) & SSIM($\uparrow$) & FID($\downarrow$) & Aesthetics($\uparrow$)\\
\midrule
Title+Image Rule-based      & 0.6446                   & 0.1484                & 47.74                             & 4.17      & 0.6512                   & 0.1634                & 46.24                             & 4.09                 \cr

Text Inversion~\cite{gal2022image} & 0.6282                   & {\ul {0.1632}}                & 42.23                             & 4.12      & 0.6345                   & 0.2474                & 43.27                             & {\ul {4.12}}\cr

PMG~\cite{PMG} & {\ul {0.5411}} & 0.1624 & {\ul {35.18}} & {\ul {4.21}} & {\ul {0.4140}} & {\ul {0.2515}} & {\ul {33.93}} & 4.11\cr

\methodnameshort    & \textbf{0.5126} & \textbf{0.1724} & \textbf{33.06} & \textbf{4.87} & \textbf{0.4018} & \textbf{0.2695} & \textbf{31.23} & \textbf{4.77} \cr
\bottomrule
\end{tabular}

\label{tab:quantitative}
\end{table*}
\begin{figure}[t]
    \centering
    \includegraphics[width=\linewidth]{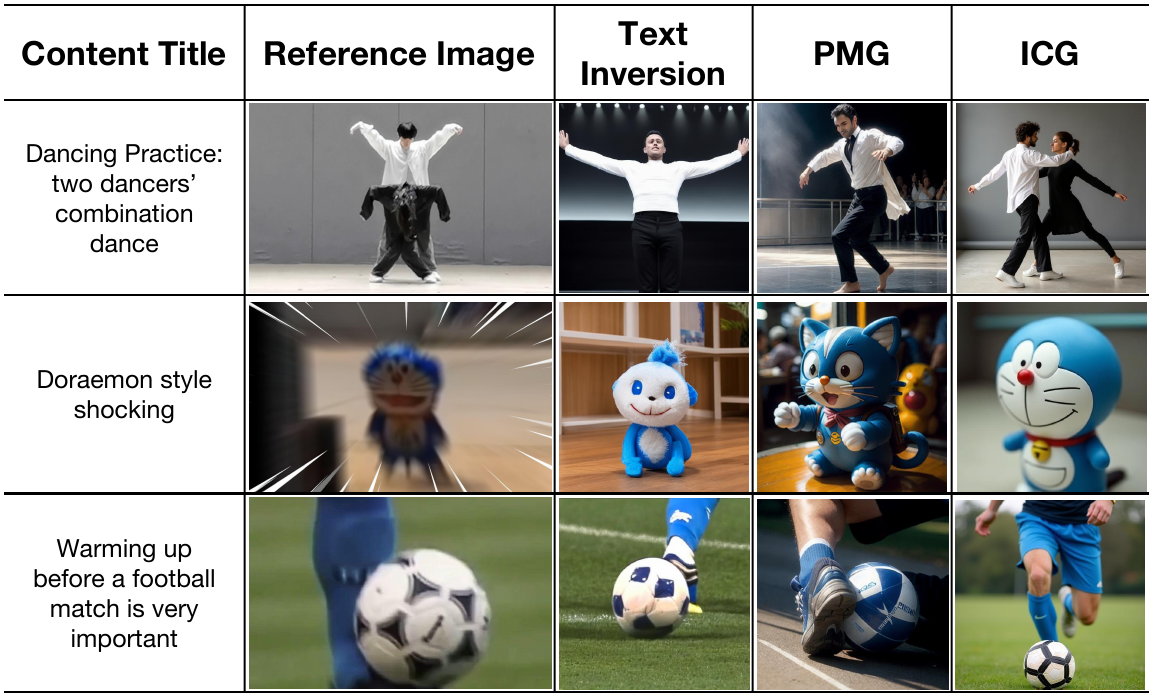}
    \caption{Qualitative comparison. Content titles, reference images and generated covers with different approaches}
    \label{fig:demo}
\end{figure}

\section{Experiments}

\subsection{Datasets and Evaluation Metrics}
We evaluate \methodnameshort on two public datasets representing short video and movie recommendation scenarios. (1) \textbf{PixelRec}\footnote{\url{https://github.com/westlake-repl/PixelRec}} is a large-scale video cover dataset; we use its 1M subset containing 0.3M covers across 22 domains, 1M user profiles, and 10M interactions, along with metadata such as clicks, likes, titles, and descriptions. (2) \textbf{MovieLens}\footnote{\url{https://grouplens.org/datasets/movielens}} includes 86K movies, 0.3M users, and 3.3M ratings, with additional user demographics and movie metadata (titles, genres, and covers). We conduct both automatic and human evaluations. For image quality, we report FID~\cite{HeuselRUNH17} and aesthetic scores using a LAION-trained predictor\footnote{\url{https://github.com/christophschuhmann/improved-aesthetic-predictor}}. For personalization, we compute LPIPS~\cite{zhang2018unreasonable} and SSIM~\cite{wang2004image} between generated and reference images. These metrics jointly assess visual appeal, fidelity, and personalization. Human evaluation further validates alignment with real-world user preferences.

\subsection{Baselines}
We compare \methodnameshort with three generative baselines: (1) \textbf{Text Inversion}~\cite{gal2022image}, which embeds user preferences into word tokens and combines them with textual prompts for diffusion-based generation; (2) \textbf{PMG}~\cite{PMG}, which transforms user-interacted and reference images into text, then extracts preference keywords via a pre-trained LLM to guide generation; and (3) a \textbf{rule-based variant of \methodnameshort}, which replaces the personalized pipeline with a vanilla Stable Diffusion model. Given only a reference image and title, the MLLM generates a drawing prompt without personalization, highlighting the benefits of end-to-end optimization and MLLM-diffusion integration.




\begin{table}[H]
\centering
\fontsize{10}{11}\selectfont
\caption{The average score of generated covers in human evaluation}
\begin{tabular}{lcc}
\toprule
                   & PixelRec &MovieLens \\ \midrule
ICG        & \textbf{2.419}  & \textbf{2.527}              \\
Title+Image Rule-based  & 1.978  &{\ul {2.041}}                \\
Text Inversion    & 1.952  &1.923                 \\
PMG    & {\ul {2.152}}   &1.994                \\
\bottomrule
\end{tabular}
\label{tab:human}
\end{table}
\subsection{Implementation details}
We use Qwen2.5VL-7B~\cite{bai2025qwen2} as the context prompt generator and adopt Stable Diffusion V1.5 or Flux for cover image generation, with adapters initialized from IP-Adapter-SD15 or IP-Adapter-Flux. During multimodal finetuning, only the adapter and projector layers are updated (meta token length = 1, projected dimension = 1024), ensuring compatibility across architectures. The full model is trained for 50,000 iterations using two 32GB GPUs, with a learning rate of $10^{-6}$ and a guidance scale of 1.0. At inference, we use the DDIM scheduler~\cite{song2020denoising} with 15 sampling steps and a guidance scale of 7.0. The personalized reward model is trained separately on PixelRec and MovieLens using 0.2M user-item pairs (80\%-10\%-10\% split), optimized with Adam (lr = $10^{-4}$) and early stopping. It consists of frozen CLIP encoders, two trainable transformer layers (768 hidden size), and fully connected heads, totaling ~20M trainable parameters.

\begin{table}
\centering
\fontsize{9}{12}\selectfont
\caption{Quantitative ablation study of multimodal generative learning stage and meta tokens using the LPIPS metric two datasets. $N$ denotes the number of multimodal tokens. The best results are in \textbf{bold} and the second-best results are \underline{underlined}.}
\begin{tabular}{lccc}
\toprule

     $N$ & Finetuning                            & PixelRec               & MovieLens        \\ \midrule
            
                 1 & \ding{55}                                        & 0.4367            & 0.5491           \\
                 2 & \ding{55}                                        & 0.4359            & 0.5482           \\
              4 & \ding{55}                                        & 0.4398            & 0.5526           \\
                  8 & \ding{55}                                        & 0.4495            & 0.5689           \\ \midrule
                  1 & \ding{51}                                        & {\ul {0.4194}}            & \textbf{0.5293}           \\
                  2 & \ding{51}                                        & \textbf{0.4168}      & {\ul {0.5315}}  \\
                  4 & \ding{51}                                        & 0.4255   & 0.5391     \\
                 8 & \ding{51}                                        & 0.4231            & 0.5412           \\
\bottomrule
\end{tabular}
\label{tab:ablation_finetune}
\end{table}
\begin{figure}[t]
    \centering
    \includegraphics[width=\linewidth]{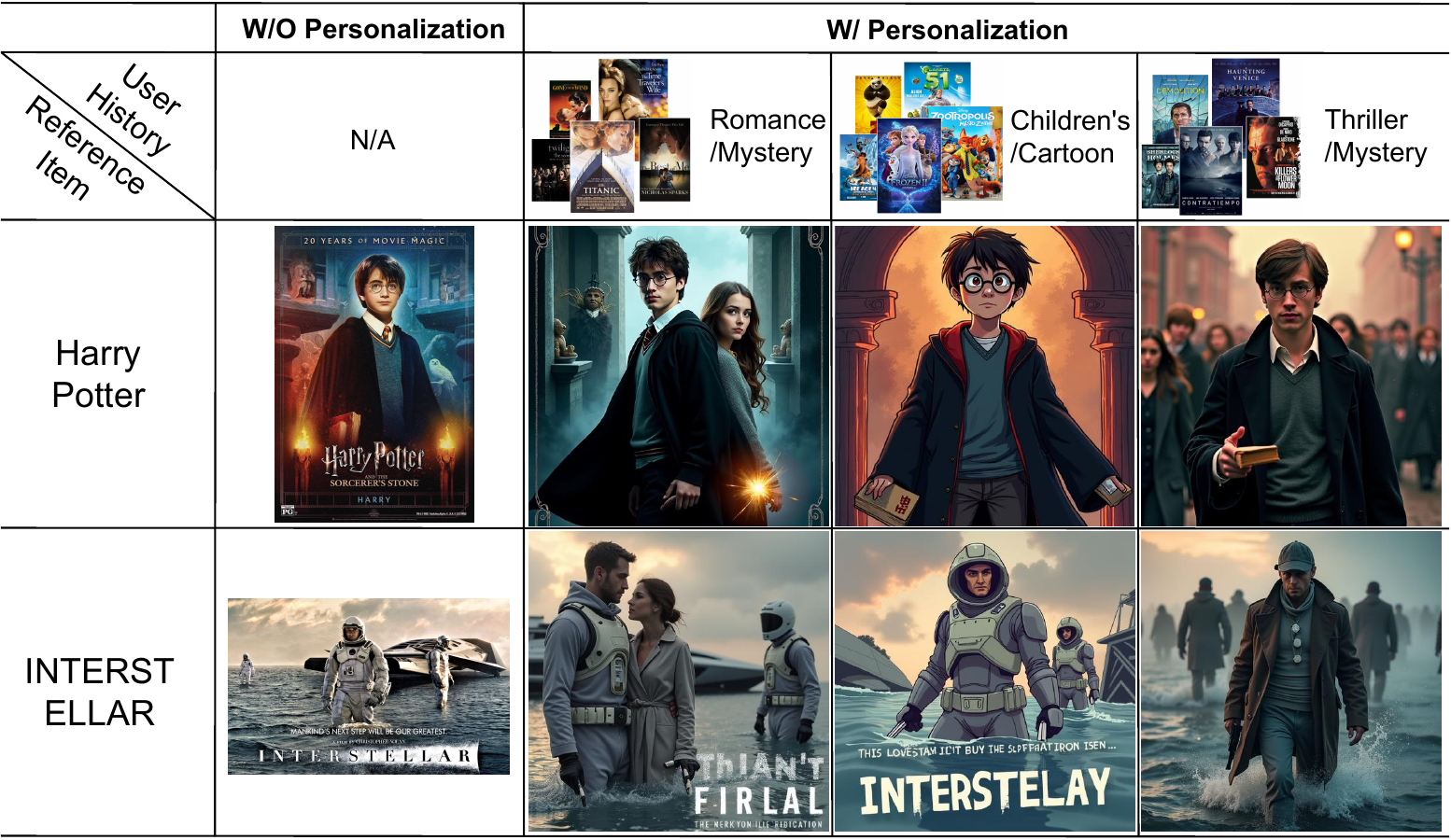}
    \caption{The effectiveness ablation of varying user conditions.}
    \label{fig:reward}
\end{figure}

\subsection{Experimental Results}
\subsubsection{\textbf{Qualitative comparison.}}
Figure~\ref{fig:demo} presents example outputs from \methodnameshort and three baselines, alongside content titles and reference images. \methodnameshort consistently achieves superior visual coherence and semantic alignment. In the first example, it accurately conveys the theme and color tone of “Dancing Practice: two dancers’ combination dance,” while baselines fail to reflect the intended meaning. The second row shows precise reconstruction of a cartoon character, whereas PMG introduces irrelevant details and Text Inversion omits key features. In the third case, it clearly depicts a warm-up scene with a football and full-body figure, effectively grounding the title, which baselines overlook. Additional results on MovieLens are discussed in later ablations.

\begin{table}
\centering
\fontsize{8}{11}\selectfont
\caption{Overall quantitative ablation study of the ICG framework. The best results are in \textbf{bold} and the second-best results are \underline{underlined}.}
\begin{tabular}{lcccc}
\toprule
Dataset             & \multicolumn{2}{c}{PixelRec}                                                        & \multicolumn{2}{c}{MovieLens}                                                  \\ \cmidrule(lr){2-3}\cmidrule(lr){4-5} 
Metric & LPIPS($\downarrow$) & FID($\downarrow$) & LPIPS($\downarrow$) & FID($\downarrow$)\\ \midrule
\methodnameshort                       & \textbf{0.5126}                                          & {\ul {33.06}}  &\textbf{0.4018}  &\textbf{31.23}                    \\
\xspace w/o Meta token                          & 0.5912       &39.24                                      & 0.5854 &37.02                     \\
\xspace w/o User feature                          & {\ul {0.5203}}                 & \textbf{32.67} & {\ul {0.4284}} & {\ul {31.43}}                    \\
\xspace w/o Both                          & 0.5893                   & 38.45 &0.5194 &36.54                     \\

\bottomrule
\end{tabular}
\label{tab:user}
\end{table}
\subsubsection{\textbf{Quantitative comparison.}}
As shown in Table~\ref{tab:quantitative}, \methodnameshort consistently outperforms all baselines on both PixelRec and MovieLens. It achieves the lowest LPIPS (0.5126, 0.4018) and FID (33.06, 31.23), and the highest SSIM (0.1724, 0.2695), indicating superior personalization, realism, and structural fidelity. While PMG performs reasonably on LPIPS and FID, it lags in aesthetics and personalization. Text Inversion and the rule-based baseline perform worst, with significantly higher LPIPS and FID. \methodnameshort also attains the highest aesthetic scores (4.87, 4.77), benefiting from joint supervision by public and personalized reward models.

\subsubsection{\textbf{Human evaluation.}}
While quantitative and qualitative results confirm the effectiveness of \methodnameshort in terms of personalization and image quality, it is necessary to further examine the user-perceived quality of the generated covers. 
To this end, we conducted a human evaluation comparing \methodnameshort with three baselines. 
A total of 100 volunteers rated 120 anonymized images (30 from each method) on a 1–3 Likert scale, with higher scores indicating stronger visual quality and alignment with user preferences. 
All images were randomly shuffled to mitigate bias. 
As shown in Table~\ref{tab:human}, \methodnameshort obtains the highest average scores, indicating stronger user appeal. 
We emphasize that this evaluation captures \emph{perceived appeal} rather than direct engagement; more detailed experiments on downstream recommendation performance are provided in the following sections.

\subsection{Ablation and Analysis}
We evaluate the impact of user feature conditions on cover generation by measuring similarity to users' historical items (personalization) and distance to the reference image (fidelity). As shown in Table~\ref{tab:user}, \methodnameshort effectively integrates user preferences, with slightly reduced reference distance in movie scenes—indicating personalization enhances alignment with original content.

\begin{figure}[t]
    \centering
    \includegraphics[width=\linewidth]{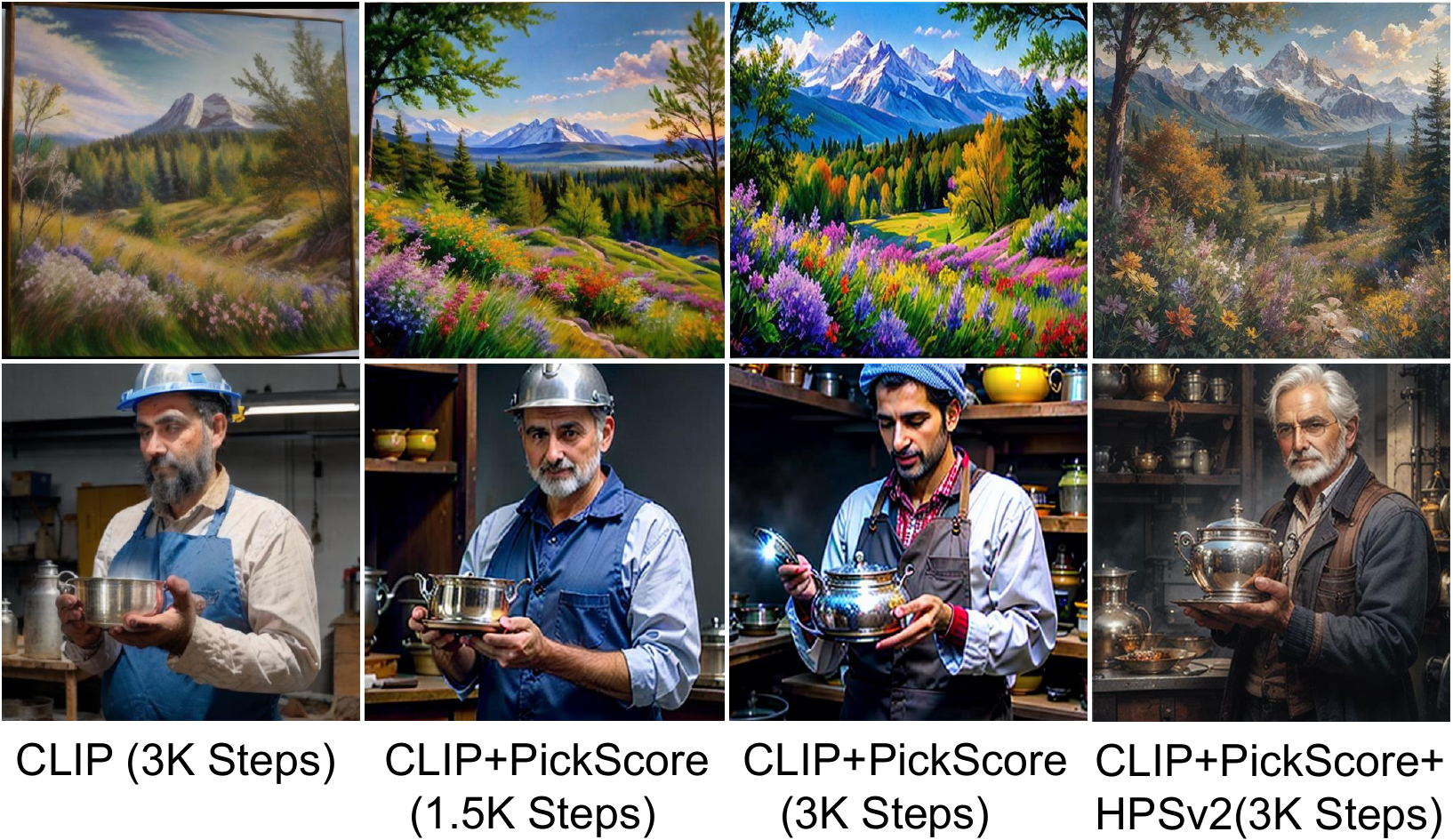}
    \caption{The effectiveness ablation of the proposed CLIP, PickScore and HPSv2 rewards.}
    \label{fig:reward_training}
\end{figure}
\begin{table}
\centering
\fontsize{7.5}{11}\selectfont
\caption{Quantitative ablation study of the reward models. The best results are in \textbf{bold} and the second-best results are \underline{underlined}.}
\begin{tabular}{lccc}
\toprule
Metric & LPIPS($\downarrow$) & FID($\downarrow$) & Aesthetics($\uparrow$) \\ \midrule
\methodnameshort                       & \textbf{0.5126}                  & \textbf{33.06}                 & \textbf{4.87}                        \\
\xspace w/o CLIP                          & 0.5504                & 35.76                              & {\ul {4.71}}        \\
\xspace w/o HPSv2+PickScore                         & {\ul {0.5413}}                 & {\ul {34.87}}                             & 4.45                        \\
\xspace w/o Personalized Reward                         & 0.5653                 & 35.81                             & 4.54                        \\
\bottomrule
\end{tabular}
\label{tab:reward}
\end{table}
We further visualize the impact of user conditions on generation. As shown in Figure~\ref{fig:reward}, for Harry Potter, the model adapts styles such as cartoon, romance, or thriller based on user preferences; for Interstellar, it integrates elements like astronauts, aliens, and oceans. These results demonstrate that \methodnameshort tailors cover styles to individual tastes while preserving core semantics.
\begin{figure}
    \centering
    \includegraphics[width=\linewidth]{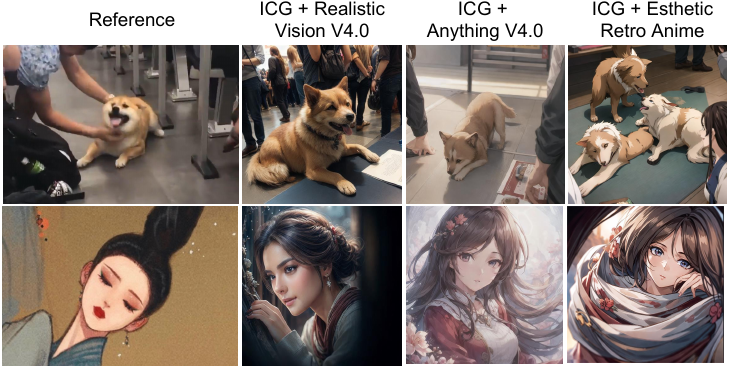}
    \caption{Generated Example Covers. Despite being trained on the base Stable Diffusion v1.5, our model can be seamlessly applied to a range of community checkpoints.}
    \label{fig:intro}
\end{figure}
\subsubsection{\textbf{Meta tokens.}}

We evaluate the impact of the multimodal generative learning stage and the number of meta tokens ($N$) on personalization using LPIPS scores on PixelRec and MovieLens. As shown in Table~\ref{tab:ablation_finetune}, finetuning notably improves performance, especially on MovieLens. For non-finetuned models, larger $N$ improves results, while finetuned models perform best at $N = 2$, with higher values degrading performance—indicating that too many tokens reduce embedding effectiveness. Table~\ref{tab:user} further shows that removing meta tokens significantly harms both personalization and image quality, underscoring their importance in capturing multimodal context.
\begin{table}[H]
\centering
\fontsize{9}{11.5}\selectfont
\caption{Personalized preference prediction accuracy on test sets of PixelRec and MovieLens under different setting}
\begin{tabular}{lcc}
\toprule
                   & PixelRec &MovieLens \\ 
                   \midrule
Personalized Reward Model        & \textbf{85.2}    &\textbf{86.2}             \\
\xspace\xspace Only image  & 53.8    &54.1              \\
\xspace\xspace Image and title & 61.3    &67.1              \\
\xspace\xspace Image and user profile & {\ul {74.6}}    &{\ul {78.3}}              \\
\xspace\xspace w/o transformers    & 70.5  &72.5                 \\
\bottomrule
\end{tabular}
\label{tab:reward_p}
\end{table}
\subsubsection{\textbf{Reward models.}}
As detailed in the Methodology, the personalized reward model is essential for enabling differentiable training in personalized cover generation. We assess its effectiveness using preference accuracy from pairwise comparisons of user-interacted items ranked by view counts (PixelRec) or ratings (MovieLens). As shown in Table~\ref{tab:reward_p}, models relying solely on image features perform poorly, while adding titles or user profiles significantly boosts accuracy. Transformer-based fusion yields further gains, underscoring the model’s ability to capture multimodal preferences. We further ablate all reward components. Figure~\ref{fig:reward_training} shows that using only CLIP similarity introduces visual distortions; adding HPSv2 improves realism but may introduce contrast bias, which PickScore helps mitigate by enhancing smoothness and sharpness. As shown in Table~\ref{tab:reward}, removing CLIP or the personalized reward notably degrades fidelity and alignment, while omitting HPSv2 or PickScore harms aesthetics. These results underscore the complementary roles of all rewards, with the personalized module being critical for modeling user-specific preferences in \methodnameshort.

\subsubsection{\textbf{Analysis of compatibility.}} 
A key advantage of our design is that the dual-path cross-attention adapter functions as a lightweight, plug-and-play module. 
Since the underlying diffusion backbone is kept frozen during training, the adapter learns to modulate attention features without altering the pretrained generative capacity of the model. 
This enables \methodnameshort to generalize beyond the base Stable Diffusion v1.5 checkpoint and seamlessly extend to custom fine-tuned variants derived from the same foundation. 
In practice, this means that once trained, our adapter can be directly inserted into diverse diffusion checkpoints without additional retraining or parameter adjustment. 

As illustrated in Figure~\ref{fig:intro}, \methodnameshort works out of the box on widely used community models from HuggingFace and CivitAi~\cite{huggingface2024civit}, including Realistic Vision V4.0~\cite{huggingface2024real}, Anything v4~\cite{huggingface2024anything}, and Esthetic Retro Anime~\cite{huggingface2024est}. 
These results highlight the universality of our module design and its practicality for real-world deployment, where model diversity and user-specific customization are essential.

\subsection{Applications in Recommendation Tasks}
\begin{table}
\centering
\fontsize{7.5}{11}\selectfont
\caption{Comparison of MMGCN's recommendation performance using different item and user image features. \textbf{Best results} are highlighted in bold, and \underline{second-best results} are underlined.}
\begin{tabular}{ccccc}
\toprule
                        & Item             & User            & Recall@10                & NDCG@10               \\ \midrule
w/o image                & \ding{55}        & \ding{55}       & 16.17\%                  & 0.0749                \\
Item               & \ding{51}        & \ding{55}       & 17.94\%                  & 0.0853                \\
Averaged-user           & \ding{51}        & Average         & {\ul {18.99\%}}            & {\ul {0.0991}}          \\
Generated-user          & \ding{51}        & Generated       & \textbf{20.21\%}         & \textbf{0.1016}       \\
\bottomrule
\end{tabular}
\label{tab:auxiliary}
\end{table}


To complement the human evaluation that primarily measured perceived appeal, we further evaluated the downstream utility of ICG-generated covers in recommendation tasks. 
Experiments were conducted on the MovieLens dataset using the multimodal recommendation model MMGCN~\cite{wei2019mmgcn}. 
We compared four input configurations: 
(1) \textit{w/o image}, using only item IDs; 
(2) \textit{Item}, using original item images; 
(3) \textit{Averaged-user}, incorporating user features from averaged images of previously interacted items; and 
(4) \textit{Generated-user}, incorporating personalized covers generated by \methodnameshort conditioned on user profiles. 
Evaluation followed standard offline recommendation metrics, including Recall@10 and NDCG@10, which are widely adopted as proxies for user engagement in large-scale recommender systems. 
Recall@10 measures the proportion of relevant items successfully retrieved within the top-10 recommendations, reflecting coverage of user interests. 
NDCG@10 (Normalized Discounted Cumulative Gain) further accounts for the ranking positions of relevant items, assigning higher importance to those appearing earlier in the list.

As shown in Table~\ref{tab:auxiliary}, adding visual features improves accuracy across all settings. 
Notably, the \textit{Generated-user} configuration achieves the strongest results, with improvements of \textbf{+2.27\% Recall@10} and \textbf{+19.1\% NDCG@10} over the best baseline. 
These results demonstrate that the personalized covers produced by \methodnameshort not only enhance visual appeal but also translate into measurable gains in recommendation effectiveness, suggesting their potential to improve downstream user engagement. 
Future work will extend this offline evaluation with online A/B testing to directly assess the impact on click-through rate and other behavioral metrics.

\section{Conclusion}
We propose \methodnameshort, a unified framework for personalized cover generation that integrates multimodal large language models (MLLMs) with diffusion models. By leveraging context and user-profile prompts, it generates outputs aligned with both item semantics and user preferences. A multi-reward learning strategy enables end-to-end optimization without the need for ground-truth labels. Experiments on two datasets demonstrate consistent improvements in image quality, semantic relevance, and personalization. As a plug-and-play module, \methodnameshort can be seamlessly integrated into existing diffusion pipelines.

\section*{Limitations}
While \methodnameshort delivers strong improvements, several limitations remain. 
First, our current framework relies on static user profile embeddings, which limits sensitivity to short-term preference shifts; integrating session-based or online user modeling is an important next step. 
Second, although multi-reward supervision enhances personalization, it introduces about 20\% additional training cost, and inference latency is roughly 1.5 seconds per image on a V100 GPU. Future work will explore model compression and accelerated diffusion backbones to enable real-time deployment. 
Finally, our evaluation of user-facing benefits is limited to offline proxies (Recall@10, NDCG@10) and human preference ratings. A direct measurement of engagement through large-scale online A/B testing remains an important avenue for future work.

\section*{Acknowledgments}
This work was supported by the National Natural Science Foundation of China under Grant No.U24A20326. We also acknowledge partial support from MindSpore (https://www.mindspore.cn), a new deep learning computing framework.

\bibliography{custom}

\appendix

\end{document}